\documentclass[letteFpaper, 10 pt, conference]{ieeeconf}  

\usepackage{amsmath}
\usepackage{multirow}
\usepackage{graphicx}
\usepackage{hyperref}
\usepackage{tikz}
\usepackage{booktabs}
\usepackage{pgfplots}
\usepackage[caption=false, font=footnotesize]{subfig}
\makeatletter
    \def\tagform@#1{\maketag@@@{\ignorespaces#1\unskip\@@italiccorr}}
    \let\orgtheequation\theequation
    \def\theequation{(\orgtheequation)}
\makeatother

\AtBeginDocument{
}

\pgfplotsset{compat=1.16}
\IEEEoverridecommandlockouts
\graphicspath{ {./images/} }

\title{\LARGE \bf
    Roughly Collected Dataset for Contact Force Sensing Catheter
}

\author{
    Seunghyuk Cho$^{*\dagger}$, Minsoo Koo$^{\dagger}$,  Dongwoo Kim, Juyoung Lee, Yeonwoo Jung, Kibyung Nam, Changmo Hwang$^\ddagger$%
    \thanks{This work was supported by the Technology Innovation Program (or Industrial Strategic Technology Development Program) (Grant No.: 20008028) funded By the Ministry of Trade, Industry \& Energy (MOTIE).}%
    \thanks{* This work was done while S.H. Cho was research intern in Asan Medical Center}%
    \thanks{$^\dagger$ Equal Contribution}%
    \thanks{$^\ddagger$ Corresponding Author}%
    \thanks{S. H. Cho, M. S. Koo, K. B. Nam and C. M. Hwang are with Asan Medical Center, Seoul, 05505, Korea. (corresponding author to provide phone: 82-2-3010-4097; fax: 82-2-3010-4182; email: {\tt\small changmo@amc.seoul.kr})}%
    \thanks{S. H. Cho, D. W. Kim, J. Y. Lee and Y. W. Jeong are with Pohang University of Science and Technology, Pohang, 37673, Korea. (email: {\tt\small shhj1998@postech.ac.kr})}%
    \thanks{M. S. Koo is with Xi’an Jiaotong University Health Science Center, Xi’an,  710061, China. (email: {\tt\small koominsoo@stu.xjtu.edu.cn})}%
    \thanks{K. B. Nam and C. M Hwang are with University of Ulsan College of Medicine, 05505, Korea. (email: {\tt\small changmo@amc.seoul.kr})}%
}

\begin{document}

\maketitle
\thispagestyle{empty}
\pagestyle{empty}

\begin{abstract}


With rise of interventional cardiology, Catheter Ablation Therapy (CAT) has established itself as a first-line solution to treat cardiac arrhythmia. Although CAT is a promising technique, cardiologist lacks vision inside the body during the procedure, which may cause serious clinical syndromes. To support accurate clinical procedure, Contact Force Sensing (CFS) system is developed to find a position of the catheter tip through the measure of contact force between catheter and heart tissue. However, the practical usability of commercialized CFS systems is not fully understood due to inaccuracy in the measurement. To support the development of more accurate system, we develop a full pipeline of CFS system with newly collected benchmark dataset through a contact force sensing catheter in simplest hardware form. Our dataset was roughly collected with human noise to increase data diversity. Through the analysis of the dataset, we identify a problem defined as Shift of Reference (SoR), which prevents accurate measurement of contact force. To overcome the problem, we conduct the contact force estimation via standard deep neural networks including for Recurrent Neural Network (RNN), Fully Convolutional Network (FCN) and Transformer. An average error in measurement for RNN, FCN and Transformer are, respectively, 2.46g, 3.03g and 3.01g. Through these studies, we try to lay a groundwork, serve a performance criteria for future CFS system research and open a publicly available dataset to public.
\end{abstract}

\section{INTRODUCTION}
Catheter Ablation Therapy (CAT) is a clinical method to treat arrhythmia, a series of irregular heartbeat caused by malfunctioning of certain tissues in heart. 
Depending on the heart rhythm, arrhythmia can be classified into tachycardia, bradycardia and fibrillation.
In the past, anti-arrhythmic drugs was the major treatment for arrhythmia.
However, CAT now becomes the first-line treatment for arrhythmia.
Comparing to the anti-arrhythmic drugs, CAT significantly reduces the chance of pulmonary vein re-connection. 
For example, Amiodarone can only temporarily relieve illness of patient~\cite{lee12}. 
Patients need to take the drug often for a long period to see effects similar to CAT.
Moreover, compared to classic surgical procedures, patients do not need to worry about inherent aesthetic, clinical problems~\cite{pmid10021475}. 
Interventional methods also offers lower risk of infection, less blood loss, and fast recovery time for patients.

\begin{figure}[t!]
  \centering
  \begin{tikzpicture}
    \draw (0, 0) node[anchor=south west] {\includegraphics[width=\linewidth]{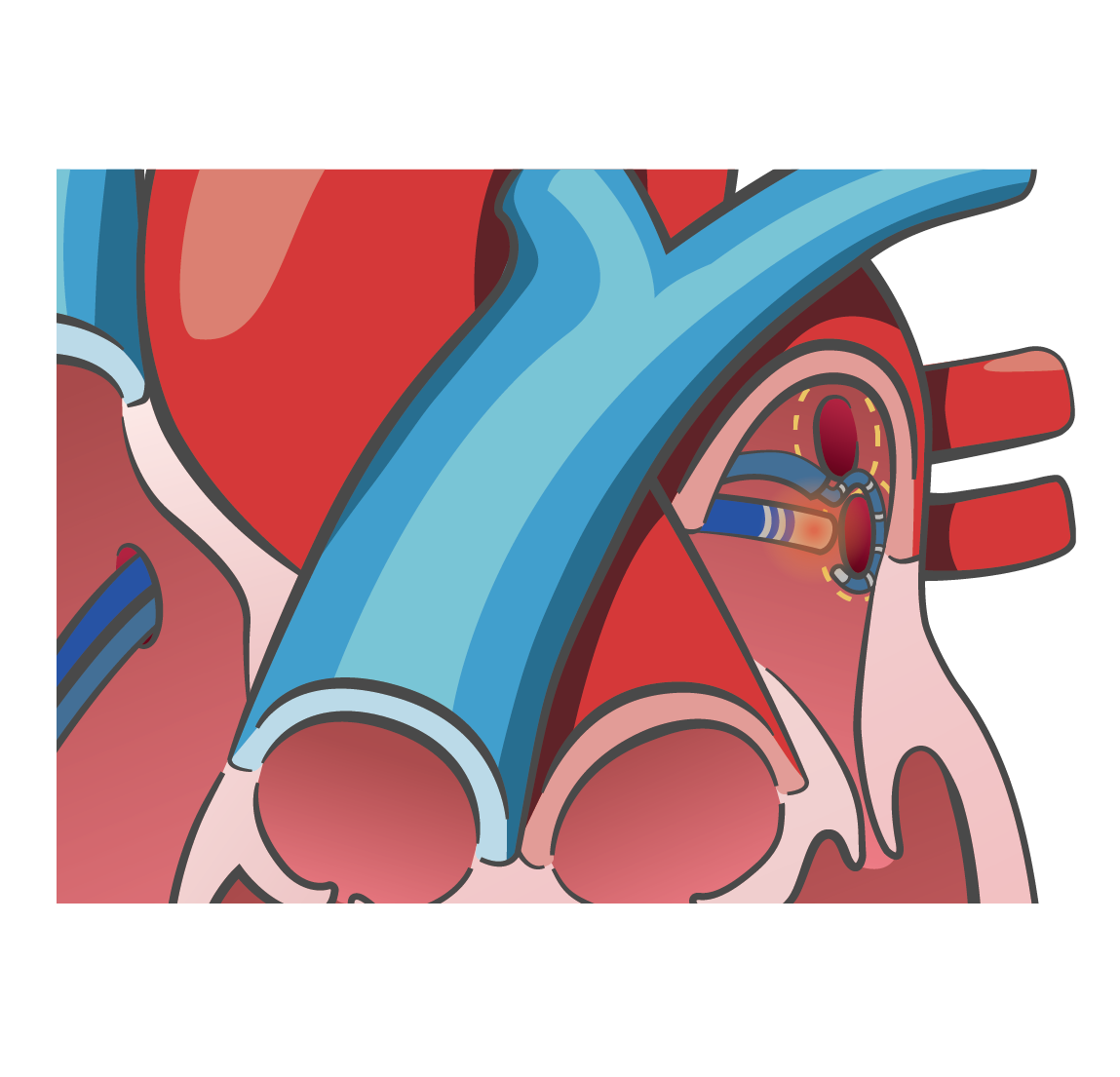}};
    \draw (2.5, 0.0) [->, thick] node {\footnotesize Normal Impulse Conduction} (2.5, 0.3)--++(-1.3, 2.2);
    \draw (7, 0.0) [->, thick] node {\footnotesize Pulmonary Vein} (7, 0.3)--++(0.8, 5);
    \draw [->, thick] (7, 0.4)--++(0.8, 4);
    \draw (4.75, 0.75) [->, thick] node {\footnotesize Left Atrium} (4.75, 1.05)--++(1.5, 2.5);
    \draw (7.3, 8.4) [->, thick] node {\footnotesize Treated Areas} (7.3, 8.1)--++(-1.04, -2.9);
    \draw [->, thick] (7.3, 8.1)--++(-0.7, -4.25);
    \draw (4.7, 7.8) [->, thick] node {\footnotesize Mapping Catheter} (4.7, 7.5)--++(-3.6, -3.99);
    \draw (2, 8.4) [->, thick] node {\footnotesize Ablation Catheter} (2, 8.1)--++(-1, -4.3);
  \end{tikzpicture}
  \caption{Illustration shows how ablation catheter is guided to dysrhythmic focus. It starts with the cardiologist inserting a 5-8 Fr wide threadlike pipe called catheter into an artery or vein through groin or arm. Then the cardiologist slowly navigate the catheter until it reaches to the left atrium. 
  If cardiologist successfully pinpoint the dysrhythmic focus and locate the catheter on it, she or he induces Radio-Frequency (RF) energy to impair the problematic cardiac tissue. The procedure is complicated and it is hard for cardiologist to know if the procedure is going well or not.}
  \label{fig:ablation}
\end{figure}
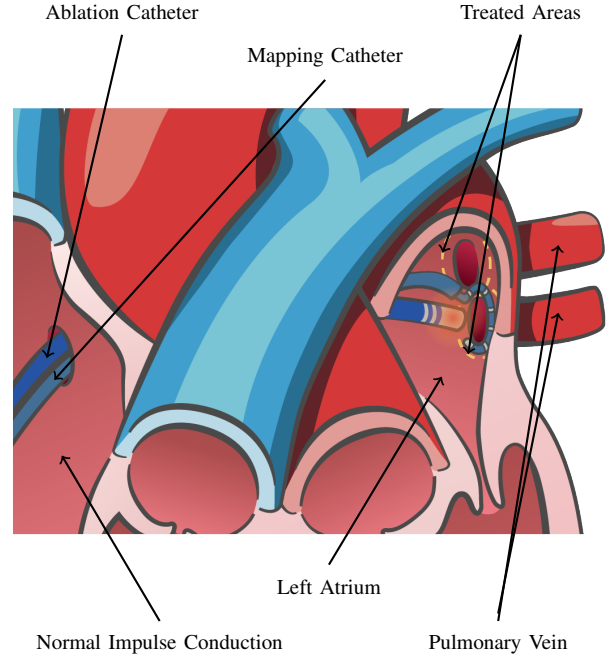

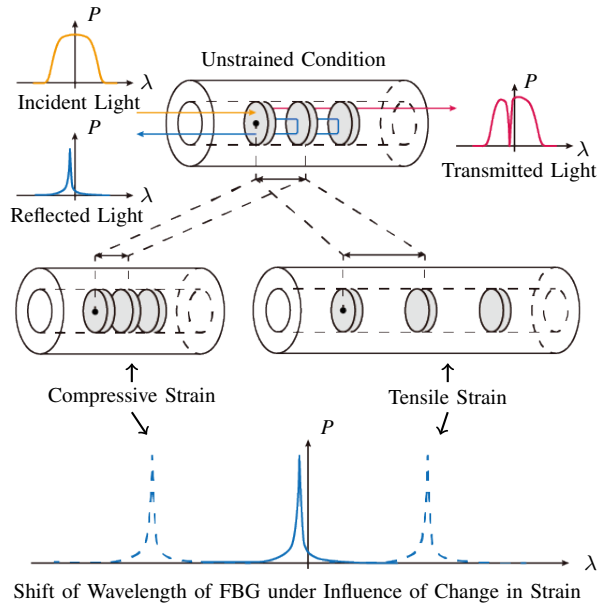
\begin{figure}[b!]
    \centering
    \begin{tikzpicture}
        \draw (0, 0) node[anchor=south west] {\includegraphics[width=\linewidth]{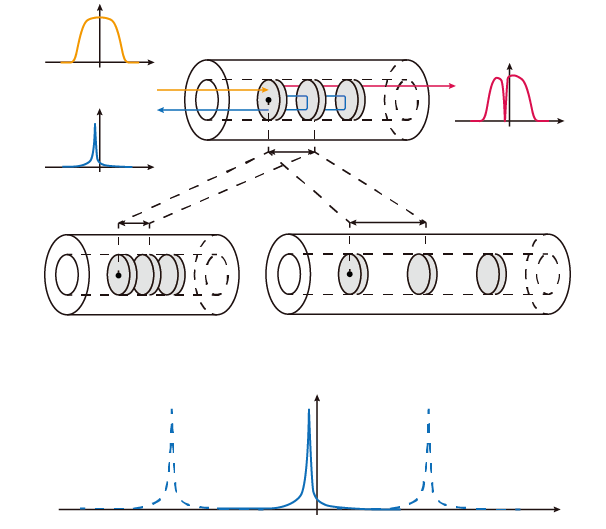}};
        \draw (2.3, 2.6) [->, thick] node {\footnotesize Compressive Strain} (2.3, 2.4)--++(0.2, -0.3);
        \draw [->, thick] (2.3, 2.8)--++(0.0, 0.25);
        \draw (6.5, 2.6) [->, thick] node {\footnotesize Tensile Strain} (6.5, 2.4)--++(-0.1, -0.3);
        \draw [->, thick] (6.5, 2.8)--++(0.0, 0.25);
        \draw (4.5, 0) node {\footnotesize Shift of Wavelength of FBG under Influence of Change in Strain};
        \draw (8.4, 0.4) node {\footnotesize $\lambda$};
        \draw (2.5, 5.3) node {\footnotesize $\lambda$};
        \draw (2.5, 6.8) node {\footnotesize $\lambda$};
        \draw (8.3, 5.9) node {\footnotesize $\lambda$};
        \draw (7.6, 6.8) node {\footnotesize \textit{P}};
        \draw (4.9, 2.2) node {\footnotesize \textit{P}};
        \draw (1.8, 7.7) node {\footnotesize \textit{P}};
        \draw (1.8, 6.2) node {\footnotesize \textit{P}};
        \draw (1.57, 5) node {\footnotesize Reflected Light};
        \draw (1.6, 6.5) node {\footnotesize Incident Light};
        \draw (7.45, 5.6) node {\footnotesize Transmitted Light};
        \draw (4.45, 7.1) node {\footnotesize Unstrained Condition};
    \end{tikzpicture}
    \caption{A FBG sensor is a phase grating inscribed to fiber optic cable. It reflects light of certain range of wavelengths and passes that of all other ranges. Optical interrogator captures the reflected light and converts it into digital figure. Temperature and strain are main factors that affect Bragg Grating wavelength. The relationship between shift of Bragg Grating wavelength and tension of FBG is proportional. If FBG receives compressive strain, Bragg Grating wavelength shifts left and vice versa.}
    \label{fig:scenario}
\end{figure}

CAT has many benefits for arrhythmia treatment but its mechanism is complex. 
\autoref{fig:ablation} illustrates the procedure of CAT for Radio-Frequency Ablation (RFA) treatment.
In CAT, even a trivial miss in procedure may lead to serious complications. Possibility of cardiac perforation, effusion and tamponade is still 0.98\% per procedure and 1.46\% per patient~\cite{ham17}.
X-Ray radiography and electrodes on catheter are developed to support the procedure, however, they are not enough for ideal procedure.
Contact Force Sensing (CFS) system is developed to solve these problems. It uses threadlike force sensor to measure the contact force of cardiac tissue - catheter contact made with Fiber Bragg Grating (FBG). 
FBG is a phase grating inscribed to the cable. 
Gating period and length, core refractive index determine whether the grating has a high or low reflectivity over a wide or narrow range of wavelengths~\cite{othonos97}. 
There are number of manuscripts about implementation of contact force sensing catheter with FBGs~\cite{back18}~\cite{back15}. 
\autoref{fig:scenario} shows the idea of how FBGs works as a force sensor.

According to~\cite{sciarra14}, it is known that CAT reduces the time of inducing RF and fluoroscopy significantly. 
Through these studies, we conclude that CFS is an indispensable system in CAT. 
The fact the system is so sensitive to accuracy mean it should be studied continuously. 
However, most of them only reports their performance without implementation and measurement details.
For example, TactiCath Quartz (TCQ) ablation catheter (Abbott Laboratories, Abbott Park, IL, USA)~\cite{bourier15}\cite{bourier16} and ThermoCool SmartTouch® Catheter (Johnson \& Johnson, New Brunswick, NJ, USA) reported their CFS precision~\cite{page12} but did not provide details of their implementation.

In our study, we develop and provide a full CFS system for accurate measurement of the contact force between cardiac tissues and catheter.
In \autoref{sec:system}, we portray what kinds of instruments were used in experiment.
In \autoref{sec:dataset}, we show the entire process of collection and preprocess. 
We collect data in environments and scenarios that are reproduced in real-world as close as possible.
Through the analysis of collected data, we identify a major problem called Shift of Reference, which hinders to accurate measurement.
In \autoref{sec:experiment}, we describe the implementation details of our CFS system.
We present two problem definitions for contact force estimation - temporary and time series.
The criteria for problem definition with a given experimental environment are also disclosed. 
In line with the problem definition, we show that deep learning models can solve it and performance of them. 
Along with it, we present a baseline that anyone could follow, showing how we had trained.
Our main contributions are the release of publicly available CFS catheter dataset and being of a guideline for future CFS implementations.

\section{CONTACT FORCE SENSING SYSTEM}
\label{sec:system}
\subsection{Hardware}

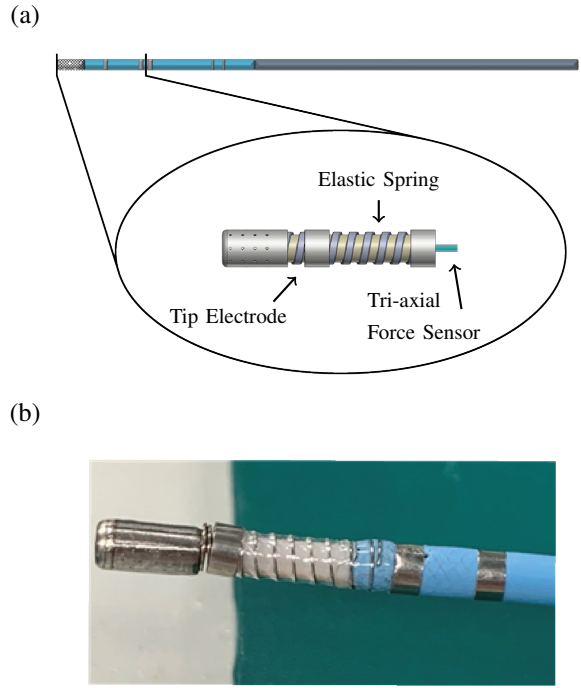
\begin{figure}[t!]
  \centering
  \begin{tikzpicture}
    \draw (0, 0) node[anchor=south west] {\includegraphics[width=\linewidth]{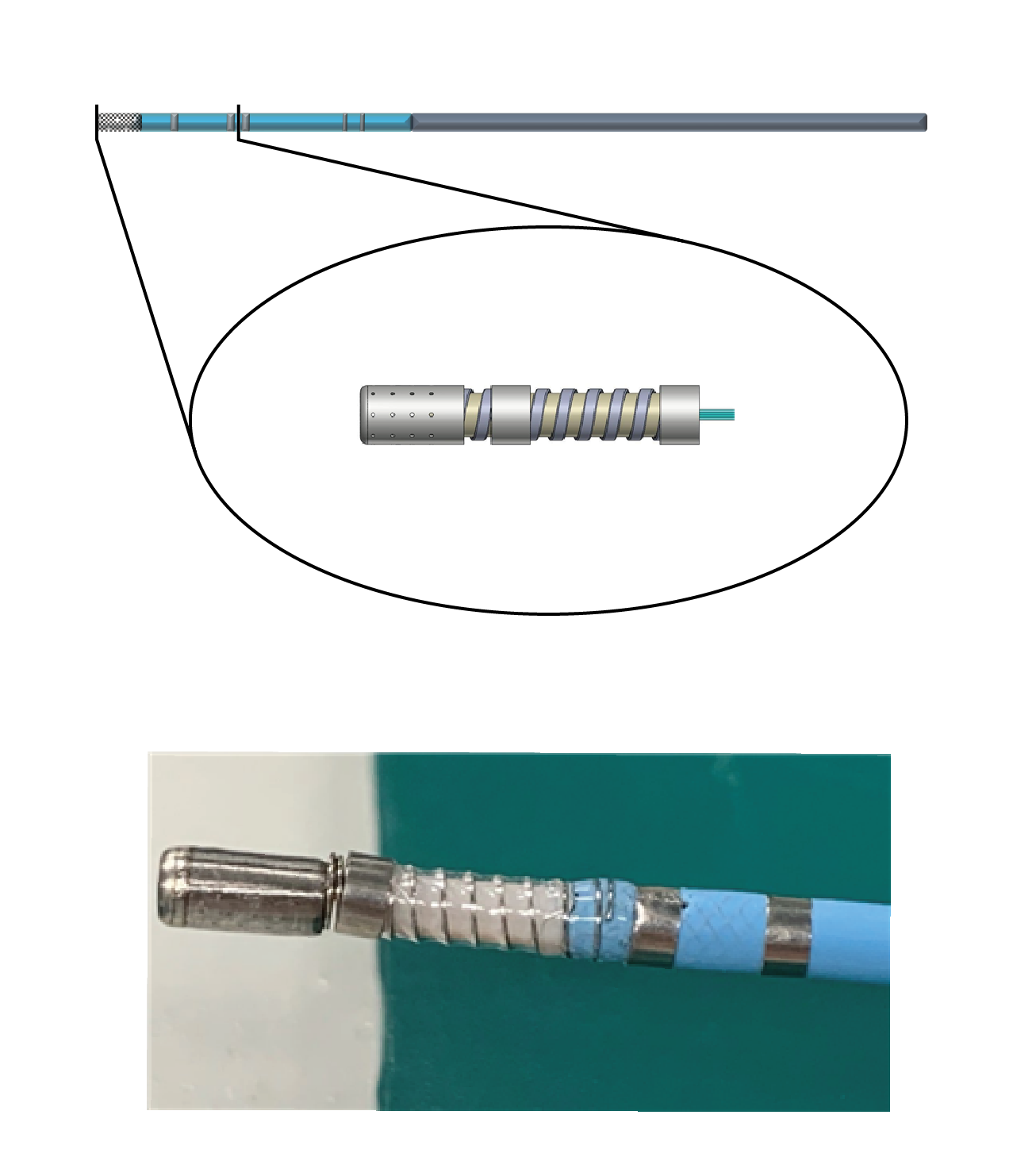}};
    \draw (0.5, 9.4) node {(a)};
    \draw (0.5, 4.1) node {(b)};
    \draw (3.2, 5.38) [->, thick] node {\footnotesize Tip Electrode} (3.8, 5.6)--++(0.3, 0.33);
    \draw (5.8, 5.4) [->, thick] node {\begin{tabular}{l} \footnotesize Tri-axial \\ \footnotesize Force Sensor \end{tabular}} (6.3, 5.5)--++(-0.1, 0.6);
    \draw (5.2, 7.2) [->, thick] node {\footnotesize Elastic Spring} (5.2, 6.95)--++(0, -0.3);
  \end{tikzpicture}
  \caption{(a) This diagram shows how tip section of catheter is built. It consists of a elastic spring, tip electrode and tri-axial force sensor built with three FBGs. We try to built a catheter with simplest form of hardware for in hope of pursuing fundamental research. FBGs and fiber optics are not allocated in a complicated, deformed shape and unnecessary components like irrigation tubes are all removed. (b) This is a photo of tip section of a catheter used in experiments.}
  \label{fig:catheter}
\end{figure}

\autoref{fig:catheter} shows the design of catheter.
CFS catheter we made is divided into multiple sections.
Core component is tri-axial force sensor. Three were used to respond to all situations of bending and to increase precision. They were molded using epoxy to heat shrink tube with same distant apart.   
The force sensor is then placed in the central lumen of catheter to remove spatial bias. 
FBG sensors used for making force sensor have (Shenzhen Lens Technology., LTD, Shenzhen, China) uniform wavelength of 1540nm, 0.5nm width, 10dB reflection rate and 6dB Side Mode Suppression Ratio (SMSR).

Optical interrogator is another indispensable device that sends light, measures wavelength of reflected light and converts the wavelength to digital signal.
Depending on the device range of wavelength, number of channels, sampling rate and precision of peak detection varies. 
We used FAZT I4W Interrogator (Femto Sensing International, Atlanta, GA, USA) for our experiment.

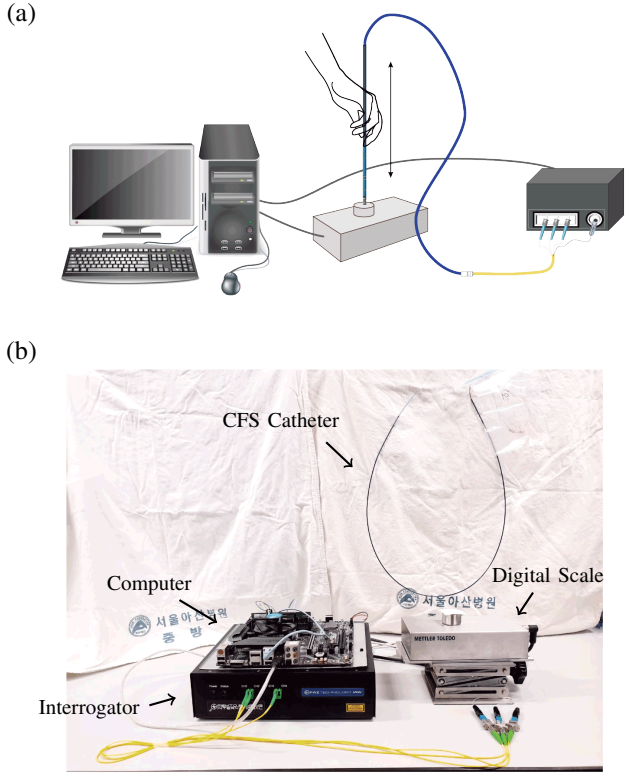
\begin{figure}[t!]
  \centering
  \begin{tikzpicture}
    \draw (0, 0) node[anchor=south west] {\includegraphics[width=\linewidth]{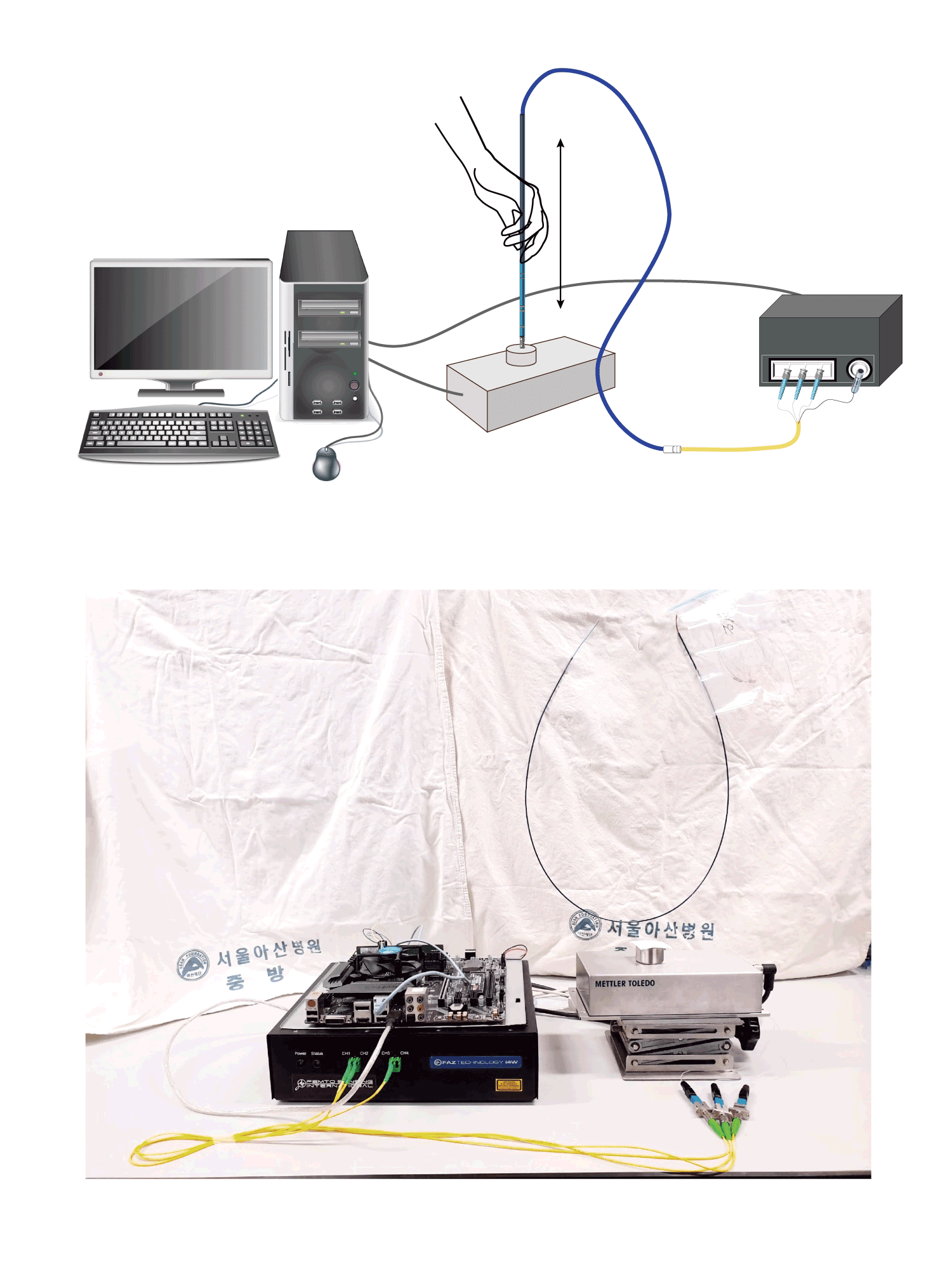}};
    \draw (0.3, 6.5) node {(b)};
    \draw (0.3, 11) node {(a)};
    \draw (3.75, 5.6) [->, thick] node {\footnotesize CFS Catheter} (4.35, 5.3)--++(0.35, -0.33);
    \draw (1.2, 1.75) [->, thick] node {\footnotesize Interrogator} (2, 1.8)--++(0.35, 0.1);
    \draw (2, 3.4) [->, thick] node {\footnotesize Computer} (2.4, 3.2)--++(0.5, -0.3);
    \draw (7.3, 3.5) [->, thick] node {\footnotesize Digital Scale} (7.1, 3.3)--++(-0.2, -0.3);
  \end{tikzpicture}
  \caption{(a) illustrates how the instruments were actually arranged during the experiment. A computer is used for synchronized and real time acquisition of data from both interrogator and digital scale. During the experiment, experimenter holds a catheter and pokes it onto the digital scale for 60 seconds. This action is repeated with different scenarios for data diversity. (b) shows the photo of instruments used in experiments. }
  \label{fig:environment}
\end{figure}

Other devices are also included like a digital scale (Mettler Toledo, Greifensee, Switzerland), a computer for recording data from interrogator and digital scale and a monitor to display the results.
\autoref{fig:environment} shows the CFS system composed with the devices mentioned above. 
Based on the environment, we will show how we collect FBG sensor signal data.

\subsection{Contact Force Estimation}

Using FBG sensor, we try to convert strain change in FBG to contact force of catheter - tissue contact.
To do this, at first, we show that wavelength shift is related to contact force.
Bragg Grating wavelength in FBG sensor is $\lambda_{B} = 2 n \Lambda$ where $n$ is refractive index and $\Lambda$ is periodic pitch size~\cite{othonos97}.
Wavelength shift from single FBG sensor can be represented as terms of strain and temperature: 
\begin{equation}
    \label{eq:shift}
    \Delta \lambda_{B} = 2(\Lambda  \frac{\delta n}{\delta l} + n \frac{\delta \Lambda}{\delta \lambda})\Delta l + 2(\Lambda  \frac{\delta n}{\delta T} + n \frac{\delta \Lambda}{\delta T})\Delta T,
\end{equation}
where $T$ is temperature and $\Delta l$ is strain.
In\autoref{eq:shift}, we see that the wavelength shift is summation of strain related term and temperature related term.
At first, to consider the effect of temperature, we attach a thermocouple wire to catheter. 
However, because the data acquisition was done in stable room temperature for whole time, there was no change of temperature.
Then we can ignore the temperature term, and rewrite\autoref{eq:shift} as 
\begin{equation}
    \label{eq:shift2}
    \frac{\Delta \lambda_{B}}{\lambda_B} = (1 - p_{e})\epsilon_{z},
\end{equation}
where $p_e$ is effective strain-optic constant, $\epsilon_z$ is axial strain and $\lambda_{B}$ is reference value of Bragg Grating wavelength. 
Equation\autoref{eq:shift2} leaves us a linear equation between axial stran $\epsilon_{z}$ and wavelength shift of FBG sensor~\cite{morey90}.

Using young's modulus, we can obtain stress with axial strain.
By definition, we can obtain contact force using stress.
So we can express contact force as term of wavelength shift:
\begin{equation}
\label{eq:ml}
\text{contact force} = f(\Delta \lambda_B).
\end{equation}
$f$ can be expressed as terms of $p_e$, strain, and etc. 
It is theoretically possible to predict contact force by measuring these physical properties.
However, there will be many noises and varies by catheters or sensors.
So we will find optimal $f$ in\autoref{eq:ml} using deep learning. 
In other words, we will show that young's modulus and other physical properties can be replaced by deep learning.

\section{CONTACT FORCE SENSING DATASET}
\label{sec:dataset}

Although CFS system is important in CAT, there is no public contact force sensing dataset for research.
To help with CFS research, we construct a CFS dataset by mapping data between interrogator and scale.
To make as similar as possible to the real situation, we randomly rotate the catheter and make it vertical to the surface.
In this section, we describe how we preprocess the collected data while the quality is ensured, and some major characteristics like Shift of Reference.

\subsection{Peak Detection}

The FBG sensor measures the intensity of each wavelength and finds and uses wavelength corresponding to peak.
The peak detection can be seen as important in FBG sensors.
Our interrogator also provides peak detection function.
However, the details of peak detection algorithm has not been released yet.
By comparing to Kernel Density Estimation (KDE) peak detection algorithm from \cite{palshikar09}, we show that interrogator's peak detection function works comparable.

The idea of KDE peak detection is to make sure that each dot is an outlier in the local context and, if correct, dismiss it as a peak.
Using Chebyshev inequality, we can state the conditions: 
Let the $i$th signal value be $x_i$. Then 
(i) $x_i > m$ where $m$ is the mean of entire signal and 
(ii) $|x_i - m| > h \cdot s$ for some suitably chosen $h > 0$, where $s$ is the standard deviation \cite{palshikar09}.
KDE peak detection uses a newly calculated value instead of writing the $x$ value as it is: 
\begin{align}
\label{eq:kde}
    S(x_i) &= H_w(N(k, i)) - H_w(N'(k, i)), \\
\label{eq:entropy}
    H_w(A) &= \Sigma_{i=1}^M(-p_w(a_i)\log(p_w(a_i))), \\
\label{eq:prob}
    p_w(a_i) &= \frac{1}{M|a_i - a_{i + w}|} \Sigma_{j=1}^M K\left(\frac{a_i - a_j}{|a_i - a_{i + w}|}\right),
\end{align}
where $N(k, i)$ is $\{x_{i - k}, \cdots, x_{i - 1}, x_{i + 1}, \cdots, x_{i + k}\}$ and $N'(k, i)$ is $\{x_{i - k}, \cdots, x_{i + k}\}$.
$N(k, i)$ is local context of $x_i$ without it while $N'(k, i)$ contains.
KDE peak detection use \autoref{eq:kde}, difference of entropy as a value with Chebyshev inequality.
Details of the calculation of entropy can be seen in \autoref{eq:entropy} and \autoref{eq:prob}
For kernel function $K$ in \autoref{eq:prob}, we used Gaussian kernel: $K(x) = \frac{1}{\sqrt{2\pi}}e^{-\frac{1}{2}x^2}$.

\begin{table}[b!]
    \centering
    \caption{Statistics of Two Different Peak Detection Algorithm on 0g Contact Force Signal Data}
    \begin{tabular}{l c c c c c c}
        \toprule
        Algorithm & \multicolumn{3}{c}{KDE} & \multicolumn{3}{c}{Femto Sensing} \\
        \midrule
        Sensor \#& 1 & 2 & 3 & 1 & 2 & 3 \\
        \midrule
        $\mu$&1539.4&1539.7&1539.8&1539.7&1539.7&1539.5\\
        $\sigma$&8.1e-2&1.3e-1&1.0e0&6.2e-3&7.8e-3&1.2e-2 \\
        \bottomrule
    \end{tabular}
    \label{table:peak detection}
\end{table}

We test both algorithms on 0g contact force signal data.
By comparing the statistic of the two algorithms, we tried to justify the Femto Sensing's algorithm.
Because we used 0g contact force data, Femto Sensing's algorithm should have similar or lower standard deviation between to KDE algorithm.
As a result of the actual experiment in \autoref{table:peak detection}, we were able to confirm that the two showed similar standard deviations. 
It has been concluded that Femto Sensing's peak detection algorithm can be used as it is.

\subsection{Preprocess}

Both of the collected data from interrogator and scale have inconsistent frequency.
Although \(\Delta t\) can be used together to solve it, this makes the problem more complicate.
Furthermore each frequency of information has a near value.
So we resample the collected data. 
Resampling is mainly used to estimate information in a specific time zone where information has not been collected.
We use cubic interpolation as a resampling method for both data.
After preprocessing, each data has a constant frequency of 1000 Hz and 10 Hz.

\subsection{Analysis}

\begin{figure}[t!]
    \centering
    \begin{tikzpicture}
        \begin{axis}[area style, width=7cm, scale only axis, height=1.5cm]
            \addplot+[green, ybar interval, mark=none] table[col sep=comma, x=idx, y=s0]{data/histogram.csv};
            \legend{sensor1}
        \end{axis}
        \begin{axis}[area style, width=7cm, scale only axis, height=1.5cm, anchor=north west, yshift=-1.0cm]
            \addplot+[pink, ybar interval, mark=none] table[col sep=comma, x=idx, y=s1]{data/histogram.csv};
            \legend{sensor2}
        \end{axis}
        \begin{axis}[area style, width=7cm, scale only axis, height=1.5cm, anchor=north west, yshift=-3.5cm, xlabel=$\Delta \lambda_B$(nm)]
            \addplot+[brown, ybar interval, mark=none] table[col sep=comma, x=idx, y=s2]{data/histogram.csv};
            \legend{sensor3}
        \end{axis}
    \end{tikzpicture}
    \caption{The above graphs are the histogram of the wavelength shift values of each FBG sensor. All three graphs are similar. It is a form in which frequency increases as it approaches zero and decreases as it moves away. Furthermore, it's almost symmetrical so that positive and negative values are distributed evenly.}
    \label{fig:histogram}
\end{figure}
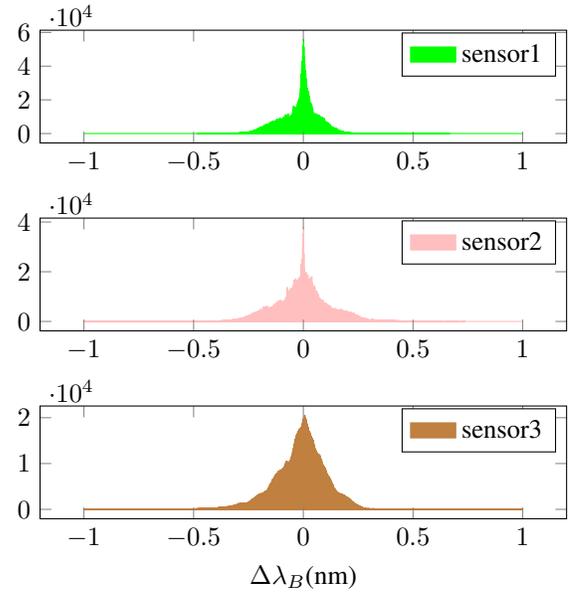

Our catheter's tip is bent while collecting data and the contact force is overly large. 
If two kinds of preconditions are satisfied and large force is exerted, two FBG sensor wavelength shift positively while the other one negatively or vice versa;
\autoref{fig:scenario} shows the details.
Two preconditions are when the guide wire bent the tip part of catheter or the angle of catheter-tissue is far from normal. 
Summing up, when the force applied is large enough to make the bent catheter even more, this kind of scenario happens. 
Functionally, this is inevitable. 
If there is not enough space in the heart chamber to allow catheter to be straight, cardiologist need to bend the catheter by one of two preconditions mentioned upper.

We validate the FBG sensor data through showing the above situation occurs evenly.
In \autoref{fig:histogram} which is the histograms of our FBG sensor data, the closer to zero, the more data there is, and the farther away the less.
This seems to be the case because most of the time catheter is not in contact.
Furthermore, all of the three FBG sensors have both positive and negative values uniformly.
Due to the randomness of rotation angle, the uniform distribution of values means that the pattern mentioned above appears evenly.
As a result, the data quality is still ensured after preprocess.

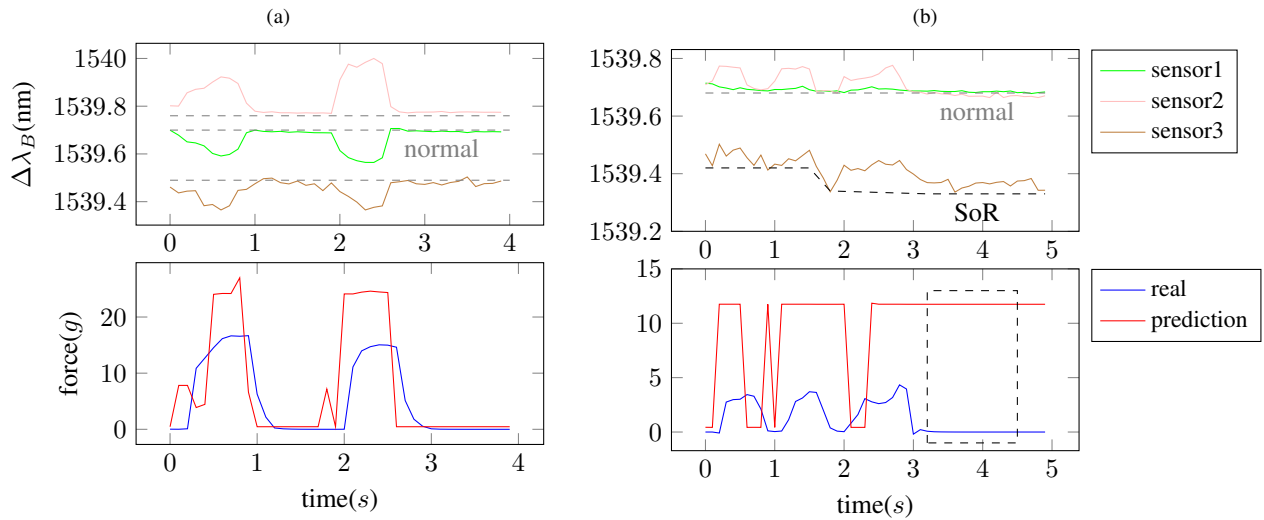
\begin{figure*}[t!]
    \centering
    \captionsetup{position=top}
    \pgfplotsset{width=7cm, height=4cm, /pgf/number format/.cd, 1000 sep={}}
    \subfloat[]{%
        \begin{tikzpicture}
            \begin{axis}[ylabel=$\Delta \lambda_B$(nm)]
                \addplot[green, mark=none] table[col sep=comma, x=time, y=s0]{data/sor_good.csv};
                \addplot[pink, mark=none] table[col sep=comma, x=time, y=s1]{data/sor_good.csv};
                \addplot[brown, mark=none] table[col sep=comma, x=time, y=s2]{data/sor_good.csv};
                \addplot[gray, mark=none, dashed] coordinates {(0,1539.7) (4,1539.7)}
                node[pos=0.8, below]{normal};
                \addplot[gray, mark=none, dashed] coordinates {(0,1539.76) (4,1539.76)};
                \addplot[gray, mark=none, dashed] coordinates {(0,1539.49) (4,1539.49)};
            \end{axis}
            \begin{axis}[anchor=north west, yshift=-0.5cm, xlabel=time($s$), ylabel=force($g$)]
                \addplot[blue, mark=none] table[col sep=comma, x=time, y=real] {data/pred_good.csv};
                \addplot[red, mark=none] table[col sep=comma, x=time, y=pred] {data/pred_good.csv};
            \end{axis}
        \end{tikzpicture}
        \label{fig:sor good}
    }\quad
    \subfloat[]{%
        \begin{tikzpicture}
            \pgfplotsset{legend style={cells={anchor=east}, legend pos=outer north east, font=\small}, legend cell align={left}}
            \begin{axis}[ymin=1539.2]
                \addplot[green, mark=none] table[col sep=comma, x=time, y=s0]{data/sor_bad.csv};
                \addplot[pink, mark=none] table[col sep=comma, x=time, y=s1]{data/sor_bad.csv};
                \addplot[brown, mark=none] table[col sep=comma, x=time, y=s2]{data/sor_bad.csv};
                \addplot[gray, mark=none, dashed] coordinates {(0,1539.68) (4.9,1539.68)}
                node[pos=0.8, below]{normal};
                \addplot[black, mark=none, dashed] coordinates {(0,1539.42) (1.5,1539.42) (1.8,1539.34) (3.2,1539.33) (4.9,1539.33)}
                node[pos=0.8, below]{SoR};
                \legend{sensor1,sensor2,sensor3}
            \end{axis}
            \begin{axis}[anchor=north west, yshift=-0.5cm, xlabel=time($s$), ymax=15]
                \addplot[blue, mark=none] table[col sep=comma, x=time, y=real] {data/pred_bad.csv};
                \addplot[red, mark=none] table[col sep=comma, x=time, y=pred] {data/pred_bad.csv};
                \draw[dashed] (3.2,-1.0) rectangle (4.5,13);
                \legend{real, prediction}
            \end{axis}
        \end{tikzpicture}
        \label{fig:sor bad}
    }\\
    \caption{The above graphs show the wavelength shift of FBG sensors according to the contact force, while the graphs below show the results of predictions of the FCN model simply trained. (a) shows that the wavelength of the FBG sensor returns to reference wavelength when the contact force reaches 0g. Furthermore, we can see that the trend of the prediction of the FCN model follows the contact force. However, in the case of sensor 3 of (b), it can be seen that the wavelength does not return to the reference even if the contact force reaches 0g. Because the wavelength didn't returned to reference, we can see that the prediction is well out of 0g. We defined the phenomenon as \textit{Shift of Reference}.}
    \label{fig:sor}
\end{figure*}

After training a simple Fully Connected Network (FCN), we find that there is an unexpected problem with the data.
\autoref{fig:sor}a shows the normal case of FBG sensor data and contact force estimation.
When the contact force reaches 0g, the wavelength of FBG sensors also returns to the reference value, even the FCN model also predicts 0g.
Furthermore, the trend of the prediction and FBG sensor data exactly follows the contact force.
However in \autoref{fig:sor}b, we can see that some of the FBG sensor doesn't return to the reference wavelength.
Due to this phenomenon, the FCN model starts to make a wrong prediction of contact force.
In \autoref{fig:sor}b, the error is almost 10g when the contact force is 0g.
We can see that even if only one sensor went wrong, it had a big impact on contact force estimation.
Furthermore, we can see that the model's prediction is about 0.4g floating.
This means that the frequency of this phenomenon is not small and effecting the whole prediction.
As a result, we decide that this phenomenon is one of the property of our data, and defined as \textit{Shift of Reference (SoR)}.

The cause of SoR is not yet clear.
Basically, the fiber optics are attached to the heat shrink tube in the catheter with either epoxy molding or UV molding.
\autoref{fig:catheter} shows this structure.
During the experiments, various forces are applied like electric, thermal force.
By hydraulic chamber or mechanic cable actuation, the force is transmitted over the entire body of catheter.
So we expect that when experiments become more frequent, the molded part is broken due to various forces, and the frequency of SoR is often generated.
We will conduct a study to verify our expectations.

\section{BENCHMARK EXPERIMENTS}
\label{sec:experiment}

Not only did we collect the data, we also confirm through benchmark experiments that the contact force sensing actually work well in our dataset.

\subsection{Problem Definition}

The simplest problem definition will be predicting the contact force for each interrogator signal. 
However, the interrogator signal is approximately 1000Hz. When screen gets updated with such high frequency, the readability will be largely degraded.
Furthermore, the frequency of interrogator and scale is different.
For these reasons, predicting with high frequency is inappropriate in real surgical situation.
Thus, prediction rate was set to be 0.1s, meaning prediction after looking at one hundred interrogator signals.

Let the \(i\) th interrogator signal \(x_i\). 
Then the \(t\) th input of the model is \(X_t = \{x_{100 \cdot t}, \cdots, x_{100 \cdot (t+1) - 1}\}\).
Using these expressions, the probability model predicts \(t\) th contact force \(y_t\).
Firstly, only looking at the \(t\) th input, \(y_t\) can be predicted maximizing \(p(y_t | X_t)\).
We think that looking at the whole inputs is necessary to figure out SoR.
So we define our second probability model as time series problem - by maximizing \(p(y_t | X_0 \cdots X_t)\).

\subsection{Models}

After we preprocess the FBG sensor data, it goes through the model and then the decoder layer.
At last, going through the activation function, the output of the decoder layer becomes the predicted contact force.
The decoder layer is a linear layer that reduce the dimension to 1.
We use leaky ReLU for the activation function.
For the intermediate model, following models are chosen.

\paragraph*{\textbf{FCN}} 
Fully Connected Network (FCN) is made by stacking linear layers with same hidden dimension.
The input goes through an encoder layer to fit dimensions before being fed to stacked layers.
It only looks at the current input and does not have any states to save the previous results.

\paragraph*{\textbf{RNN}} 
Recurrent Neural Network (RNN) is commonly used in time series.
RNN has a state that saves previous results.
So it gives same effect as looking at the whole inputs.
There are three kinds of RNN - vanilla RNN, LSTM, and GRU.
GRU \cite{cho2014learning} is chosen for the RNN cell in our experiment.

\paragraph*{\textbf{Transformer}} 
Transformer is first introduced in Natural Language Processing (NLP) problems~\cite{vaswani2017attention}.
It can be also used for time series problems by looking at the whole inputs.
Furthermore our problem has constant frequency, we can directly use the transformer.
We use the decoder part of transformer.
It means that predicting \(t\) th contact force, it is not influenced by inputs after \(t\) th one.

\subsection{Training}
Huber loss~\cite{huber1964} is chosen for the loss function while training, and Adam optimizer is used for the optimizer without any weight or learning rate decay.
For learning rate, most of the models is trained by 1e-3.
However, some huge models, such as  RNN with 8 layers and Transformer with 4, 8 layers, are hard to optimize with learning rate 1e-3.
These three models are trained with learning rate 1e-5 and successfully converged.

\autoref{table:performance} shows the performance of predicting test data with Mean Absolute Error (MAE) performance metric.
To find the optimal model, we first fix the hidden dimension and then find the optimal number of layers.
All of the model's hidden dimension is fixed to 256.
FCN model shows best performance with 2 layers, while RNN, Transformer models show best performance with 4 layers.
After finding the optimal number of layers, we fix the number of layers and then vary the number of hidden dimension.
For Transformer, we use 12 attention heads throughout the experiments.

\begin{table}[b!]
\centering
\caption{Performance on Contact Force Dataset}
\begin{tabular}{l c c c}
\toprule
\textbf{Model} & \textbf{FCN} & \textbf{RNN} & \textbf{Transformer} \\
\midrule
\# of layers\\
1 & 3.36g & 2.76g & 3.14g \\
2 & 3.12g & 2.81g & 3.14g \\
4 & 3.21g & 2.60g & 3.10g \\
8 & 3.14g & 3.40g & 3.53g \\
16 & 3.24g & - & - \\
\midrule
\# of hidden dimension \\
64 & \textbf{3.03g} & \textbf{2.46g} & 3.08g \\
128 & 3.11g & 2.81g & 3.15g \\
256  & 3.12g & 2.60g & 3.10g \\
512 & 3.10g & 2.82g & \textbf{3.01g} \\
1024 & 3.09g & 2.64g & 3.06g \\
\bottomrule
\end{tabular}
\label{table:performance}
\end{table}
\begin{table}[b!]
\centering
\caption{Inference time of Models on V100 GPU}
\begin{tabular}{l c c}
\toprule
\textbf{Model} & \textbf{FCN} & \textbf{RNN} \\
\midrule
Average Inference time & 0.759ms & 1.380ms \\
\bottomrule
\end{tabular}
\label{table:inference}
\end{table}

\subsection{Results}

RNN model with 4 layers and hidden dimension 64 achieved state-of-the-art for our dataset.
For FCN model, 2 layers and hidden dimension 64 showed the best performance.
For Transformer model, 4 layers and hidden dimension 512 showed the best performance.
Performance of all 3 models did not improve because they were big.
Even FCN and RNN showed the best performance when the hidden dimension was 64, the smallest value in our experiment.
It showed the possibility that the problem would not be so difficult and can be figure out by simple models.

We expected that models which look the problem as time series will gain better performance with solving the SoR problem.
RNN, one of our time series model, met our expectations.
It showed a huge gap of 0.5g compared to FCN and Transformer model.
However, another time series model did not meet expectation.
In \autoref{table:performance}, the performance of Transformer model is similar to FCN model.
This showed that time series models do not necessarily show high performance.
The cause of performance decline on Transformer model will be our next study.

We also measured the inference time of models.
Transformer model was much bigger than FCN model with similar accuracy.
So we only measured the inference time for FCN and RNN model.
For each model, we used the optimal ones.
Since catheters can be used on equipment equipped with GPUs, the inference time was checked after loading the model on V100 GPU.
\autoref{table:inference} shows the results.
According to our problem definition, we have to predict contact force at interval of 0.1s.
Both models meet this requirements.

\section{CONCLUSIONS}

Through these studies, we provide a foundation of CFS system.
We present a full pipeline from dealing with raw FBG sensor data to contact force estimation.
Furthermore, we release a contact force dataset generated by catheter that could serve as the basis for future research.
While providing the benchmark, we show the potential of deep learning for contact force sensing.

For future research, more techniques can be applied on our dataset.
Basically, canonical filters are available in signal processing tasks, but they have not been applied yet~\cite{vaseghi2008advanced}.
On the benchmark experiments, we only try fundamental models: FCN, RNN, and Transformer.
Since the data has been validated, more complex architectures are applicable now.

For more in-depth research, there are points to supplement in datasets.
Since the actual dataset was carried out at constant temperature during the production process, the noise by temperature was negligible.
However, temperature changes are frequent in RFA, cryoablation~\cite{Skanes2002}.
Furthermore this dataset only considered cases where the catheter is in vertical contact with the ground.
In actual surgical situations, contact will be made from various angles.

\addtolength{\textheight}{-12cm}

\bibliographystyle{IEEEtran}
\bibliography{IEEEabrv,mybibfile}

\end{document}